\documentclass[nohyperref]{article}

\usepackage{microtype}
\usepackage{graphicx}
\usepackage{subfigure}
\usepackage{booktabs} 
\usepackage{color}

\usepackage{hyperref}

\usepackage[accepted]{icml2022}

\usepackage{amsmath}
\usepackage{amssymb}
\usepackage{mathtools}
\usepackage{amsthm}

\usepackage{multirow}
\usepackage[T1]{fontenc}

\usepackage[capitalize,noabbrev]{cleveref}

\theoremstyle{plain}

\theoremstyle{definition}

\theoremstyle{remark}

\usepackage[textsize=tiny]{todonotes}
\newcommand{\method}{BCL }
\newcommand{\methodnospace}{BCL}

\icmltitlerunning{Contrastive Learning with Boosted Memorization}

\begin{document}

\twocolumn[
\icmltitle{Contrastive Learning with Boosted Memorization}




\begin{icmlauthorlist}
\icmlauthor{Zhihan Zhou}{sjtu}
\icmlauthor{Jiangchao Yao}{sjtu}
\icmlauthor{Yanfeng Wang}{sjtu,shanghai}
\icmlauthor{Bo Han}{hkbu}
\icmlauthor{Ya Zhang}{sjtu,shanghai}
\end{icmlauthorlist}

\icmlaffiliation{sjtu}{Cooperative Medianet Innovation Center, Shanghai Jiao Tong University }
\icmlaffiliation{shanghai}{Shanghai AI Laboratory}
\icmlaffiliation{hkbu}{Department of Computer Science,
Hong Kong Baptist University}

\icmlcorrespondingauthor{Jiangchao Yao}{Sunarker@sjtu.edu.cn}
\icmlcorrespondingauthor{Yanfeng Wang}{wangyanfeng@sjtu.edu.cn}

\icmlkeywords{Machine Learning, ICML}

\vskip 0.3in
]



\printAffiliationsAndNotice{}  

\begin{abstract}
Self-supervised learning has achieved a great success in the representation learning of visual and textual data. However, the current methods are mainly validated on the well-curated datasets, which do not exhibit the real-world long-tailed distribution. Recent attempts to consider self-supervised long-tailed learning are made by rebalancing in the \emph{loss perspective} or the \emph{model perspective}, resembling the paradigms in the supervised long-tailed learning. Nevertheless, without the aid of labels, these explorations have not shown the expected significant promise due to the limitation in tail sample discovery or the heuristic structure design. Different from previous works, we explore this direction from an alternative perspective, \textit{i.e.,} the \emph{data perspective}, and propose a novel Boosted Contrastive Learning (BCL) method. Specifically, BCL leverages the memorization effect of deep neural networks to automatically drive the information discrepancy of the sample views in contrastive learning, which is more efficient to enhance the long-tailed learning in the label-unaware context. Extensive experiments on a range of benchmark datasets demonstrate the effectiveness of BCL over several state-of-the-art methods. Our code is available at \href{https://github.com/MediaBrain-SJTU/BCL}{https://github.com/MediaBrain-SJTU/BCL}.

\end{abstract}

\section{Introduction}
Self-supervised learning~\citep{doersch2015unsupervised,wang2015unsupervised} that learns the robust representation for downstream tasks have achieved a significant success in the area of computer vision~\citep{chen2020simple,he2020momentum} and natural language processing~\citep{lan2019albert,brown2020language}. Nevertheless, previous studies are mainly conducted on the well-curated datasets like ImageNet~\citep{deng2009imagenet}, which is usually balanced among categorizes. In comparison, the real-world natural sources usually follow a long-tailed, even heavy-tailed distributions~\citep{reed2001pareto} that is challenging to learn for the current machine learning methods. Specially, recent attempts~\citep{jiang2021self} have shown that self-supervised learning under long-tailed distribution still requires more explorations to achieve the satisfying performance~\citep{van2018inaturalist}.

Existing works for self-supervised long-tailed learning are mainly from the \emph{loss perspective} or the \emph{model perspective}. The former relies on the loss reweighting, \textit{e.g.,} the focal loss in hard example mining~\citep{lin2017focal} or SAM by means of the sharpness of the loss surface~\citep{liu2021self}, to draw more attention on tail samples during training. However, the effectiveness of these methods is sensitive to and limited by the accuracy of the tail sample discovery. The latter mainly resorts to the specific model design like the divide-and-contrast ensemble~\citep{tian2021divide} or self-damaged-contrast via pruning~\citep{jiang2021self} to make the model better capture the semantics of the tail samples. These designs require the empirical heuristic and are usually black-box to understand the potential working dynamics for the further improvement~\citep{zhang2021deep}.

\begin{figure}[!t]
	\centering
	\includegraphics[width=0.241\textwidth]{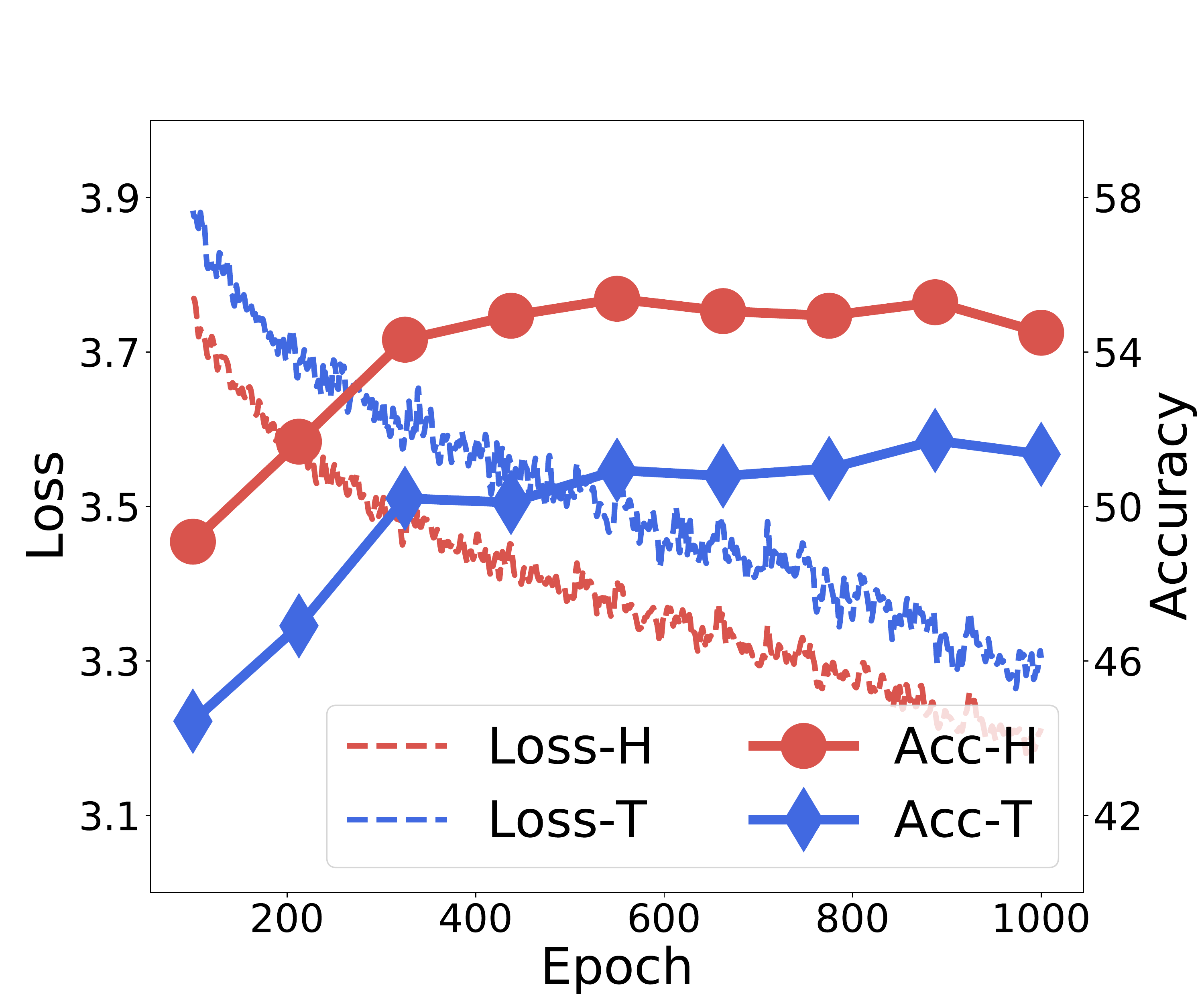}
	\hspace{-2mm}
	\includegraphics[width=0.241\textwidth]{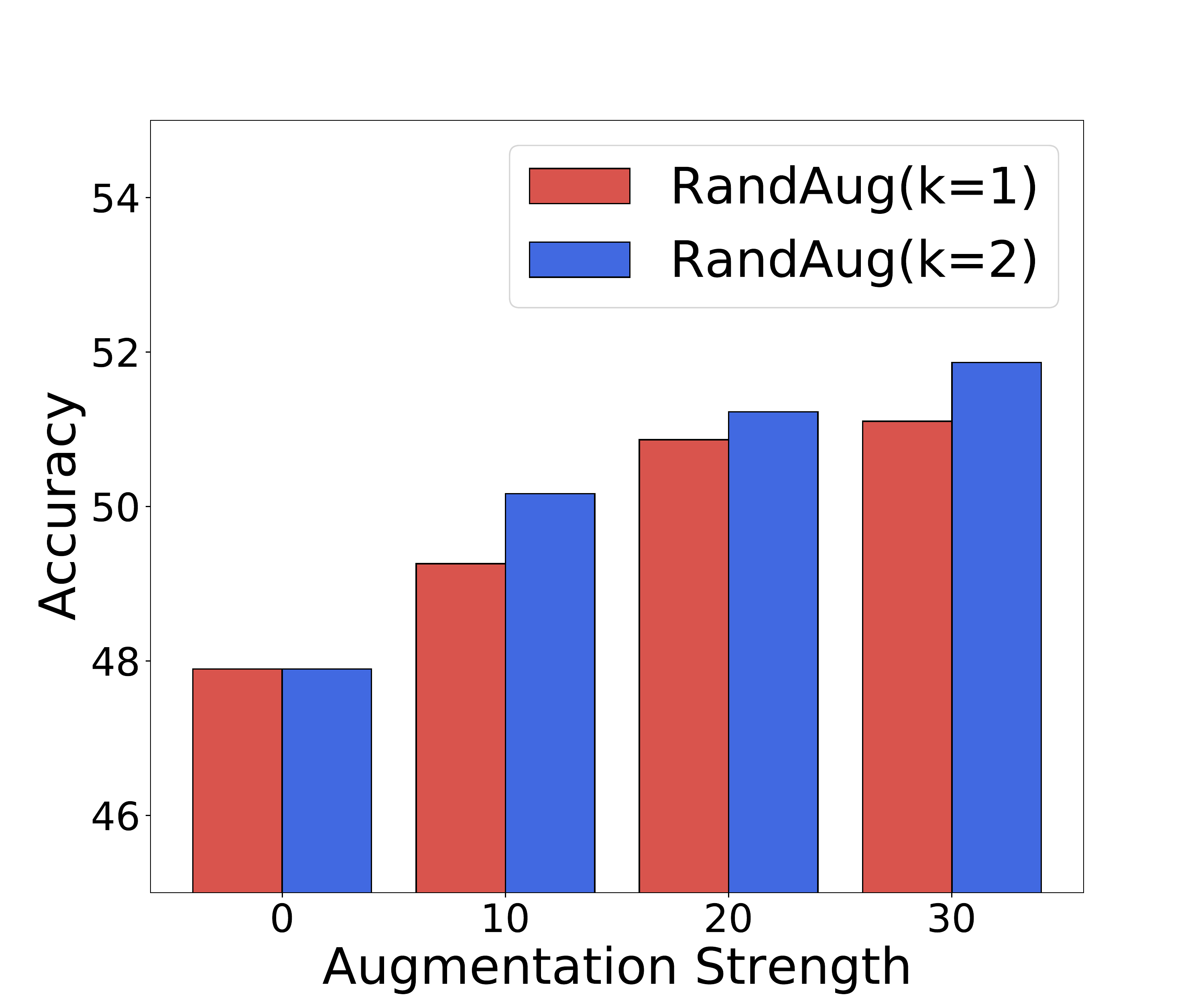} 
    \caption{(Left) Test accuracy and loss  of head and tail classes during the training stage on CIFAT-100-LT. (Right) Test accuracy of tail classes when deploying different strength of RandAugment on tail classes on CIFAT-100-LT. $k$ is a hyper-parameter controlling the amount of augmentations used in RandAugment.}
    \label{fig:intuition}
\end{figure}

In this paper, we propose to study the self-supervised long-tailed learning in the \emph{data perspective}. Our framework is motivated by the memorization effect~\citep{zhang2017understanding,arpit2017closer,feldman2020does} of deep neural networks on data, where the easy patterns are usually memorized prior to the hard patterns. As shown in the left panel of Figure~\ref{fig:intuition}, the memorization effect still holds under long-tailed datasets, where the loss and accuracy of the tail samples consistently fall behind those of head samples. This inspires us to approximately distinguish the head and tail samples by analyzing the memorization effect. Another important motivation is except the loss reweighting or model re-design, the data augmentation is very effective in self-supervised long-tailed learning to achieve the improvement by introducing the information discrepancy of two views~\citep{tian2020makes}. As illustrated in the right panel of Figure~\ref{fig:intuition}, we can see that the heavier augmentation consistently boosts the performance of the treatment tail samples. Besides, the data augmentation does not directly modify the loss or the model structure and thus is more robust to the noisy tail discovery.

On basis of the aforementioned observations in Figure~\ref{fig:intuition}, we introduce a novel Boosted Contrastive Learning method from the data perspective. Concretely, we propose a momentum loss to capture the clues from the memorization effect of DNNs to anchor the mostly possible tail samples. Then, the momentum loss is used to drive an instance-wise augmentation by constructing different information discrepancy for head and tail samples. In an end-to-end manner, \method maintains the learning of head samples, meanwhile enhances the learning of hard-to-memorize tail samples. 

\paragraph{Main Contributions}
\begin{itemize}
    \item Different from previous works in the loss and model perspectives, we are the first to explore self-supervised long-tailed learning from the data perspective, which leverages the DNN memorization effect on data and the augmentation efficiency in self-supervised learning.

    \item We propose a Boosted Contrastive Learning method, which builds a momentum loss to capture clues from the memorization effect and drive the instance-wise augmentation to dynamically maintain the learning of head samples and enhance the learning of tail samples.
    
    \item The proposed \method is orthogonal to the current self-supervised methods on long-tailed data. Extensive experiments on a range of benchmark datasets demonstrate the superior performance of \methodnospace.

\end{itemize}

\section{Related Works}

\textbf{Supervised Long-tailed Learning.}
Recent works~\citep{yang2020rethinking,kang2020exploring} start to boost the long-tailed recognition via the lens of representation learning~\cite{zheng2019understanding}. \citet{kang2019decoupling} proposed to disentangle representation and classification learning in a two-stage training scheme and empirically observed that the instance-balanced sampling performs best for the first stage, which attracts more attention to representation learning in long-tailed recognition. \citet{yang2020rethinking} theoretically investigated the necessity of the label information for long-tailed data and showed the promise of self-supervised pre-training stage on long-tailed recognition. Motivated by these findings, \citet{kang2020exploring} first leveraged supervised contrastive learning paradigm for long-tailed recognition and claimed that the learned feature space is more balanced compared with the supervised learning. \citet{cui2021parametric} theoretically showed that supervised contrastive learning still suffers from the bias from the head classes under imbalanced data. They proposed a parametric class-wise learnable center to rebalance the contrastive loss across different class cardinality. The concurrent work~\citep{li2021targeted} proposed a uniform class center assignment strategy to force a balanced feature space. 

\textbf{Self-supervised Long-tailed Learning.}
In self-supervised learning area, several works~\citep{chen2020simple, he2020momentum, chen2021exploring} mainly target to the curated and balanced dataset and naturally build the uniformity assumption. For example, \citet{wang2020understanding} concluded that one key property of contrastive learning is to learn a uniform feature space by information maximization.  \citet{caron2020unsupervised} assumed that all the samples are distributed uniformly at the prototype level and operated the fast Sinkhorn-Knopp algorithm~\citep{cuturi2013sinkhorn} for the uniform online clustering.  However, 
it may cause performance degeneration to model the real-world distribution in a uniform way as the practical data generally follows a skewed distribution\citep{reed2001pareto}. 

There exists a few attempts~\cite{liu2021self,jiang2021self,zheng2021contrastive} towards self-supervised long-tailed learning, which can be divide into two categories: \emph{loss-based} or \emph{model-based} methods. A classical solution in the first category, \emph{i.e.}, the focal loss\citep{lin2017focal}, relies on the individual sample difficulty to rebalance the learning. Recently, \citet{liu2021self} proposed a sharpness regularization on loss 
surface to enhance model generalization. From the model perspective, \citet{jiang2021self} assumed tail samples to be easily forgotten and designed a asymmetric network with a pruned branch to identify the tail classes. An alternative~\citep{tian2021divide} targeted at the uncurated data faces the similar challenges in long-tailed recognition. They proposed a multi-expert framework to extract the more fine-grained features in the separated clusters. Different from these works, we explores the benefit of the data perspective for the self-supervised long-tailed representation learning.

\textbf{Memorization Effect.}
The definition on the memorization effect of DNNs can trace back to the generalization study on noisy data~\citep{zhang2017understanding,arpit2017closer}. These findings shed lights on a stream of loss-aware studies towards noisy representation learning~\citep{jiang2018mentornet,ren2018learning,han2018co}. Specifically, they regard the small-loss samples as clean samples and then employ the sample selection or loss reweighting. For example, co-teaching~\citep{han2018co,yu2019does} selects the small-loss samples and discards high-loss samples in the training stage. Meanwhile, \citet{ren2018learning} proposed a meta-learning framework to assign different weights to the training samples according to the loss value. 

Recently, \citet{feldman2020does} extended the memorization effect of deep neural networks towards the long-tailed samples. They concluded that the memorization of DNNs is necessary for the rare and atypical instances and proposed a memorization measurement. Specifically, the memorization score are defined as the drop in the prediction accuracy for each sample in the training dataset when removing the respective sample. However, the computational cost of estimating this memorization score is expensive. The subsequent work~\citep{jiang2020characterizing} explored some more efficient proxies to alternate the hold-out estimator. In particular, a learning speed based proxy have shown the positive correlation with the memorization score, which is in consistency with the observation of the memorization effect in~\citep{feldman2020does}. Different from these explorations that 
require labels available, our methods conversely focus on the annotation-free long-tailed sample discovery.

\section{Methods}

\subsection{Preliminary}
In this section, we give the basic notations of contrastive learning that our method builds on. Generally, the classical contrastive learning~\citep{chen2020simple}, termed as SimCLR, is defined as follows,
\begin{equation}\label{eq:cl}
\mathcal{L}_{\mathrm{CL}}=\frac{1}{N} \sum_{i=1}^{N}-\log \frac{\exp \left(\frac{f(x_{i})^\top f(x_{i}^{+})}{\tau}\right)}{\sum_{x_{i}' \in X^{-}\cup \{x_i^+\} } \exp \left(\frac{f(x_{i})^\top f(x_{i}')}{\tau}\right)}
\end{equation}
where $\left(x_i, x_{i}^{+}\right)$ is the positive sample sample pair and $X^{-}$ is the negative sample set of $x$, $\tau$ is the temperature and $f(\cdot)$ is the encoder function. In practical, $x_i$ and $x_i^+$ are two views of one example, while $x_{i}'\in X^-$ is the view of other samples. Contrastive learning is to learn a representation that is invariant to itself in the small perturbation but keeps the variance among different samples.

\subsection{Motivation}
Deep supervised long-tailed learning has made great progresses in the last ten years~\citep{zhang2021deep} to handle the real-world data distributions. Nevertheless, previous works mainly focus on the supervised learning case, namely the labels of natural sources must be available, while only few works~\citep{jiang2021self,liu2021self} pay attention to the study of such a skew distribution under the self-supervised learning scenario. Compared to the supervised learning, long-tailed learning without labels is more practical and important, since in a range of cases, \textit{e.g.,} the large-scale datasets, it is expensive to collect the annotation of each sample. Concomitantly, this task is more challenging, since most of previous works build on top of the explicit label partition of head and tail samples.

Without labels, previous self-supervised learning study in this direction leverages the implicit balancing from the loss perspective~\citep{lin2017focal,liu2021self} or the model perspective~\citep{jiang2021self} to enhance the learning law on tail samples.  Different from these works, \method explicitly trace the memorization effect via a learning speed scope based on theoretical and empirical findings~\citep{feldman2020does,jiang2020characterizing} in the context of supervised image classification. The definition~\citep{feldman2020does} that describes how models memorize the patterns of the individual sample during the training is given as follows:

\begin{equation}
    \footnotesize
    \operatorname{mem}(\mathcal{A}, S, i):=
    \underset{h \leftarrow \mathcal{A}(S)}{\operatorname{Pr}}\left[h\left(x_{i}\right)=y_{i}\right]-\underset{h \leftarrow \mathcal{A}(S ^{\backslash i})}{\operatorname{Pr}}\left[h\left(x_{i}\right)=y_{i}\right] 
\end{equation}
where $\mathcal{A}$ denotes the training algorithm and $S ^{\backslash i}$ denotes removing the sample point $(x_i, y_i)$ from the data collection $S$. Unfortunately, the hold-out retraining metric is computationally expensive and only limited to the supervised learning. Inspired by the learning speed proxy explored in the subsequent work~\citep{jiang2020characterizing}, we first extend the memorization estimation to the self-supervised learning task. Specifically,
we propose the momentum loss to characterize the learning speed of individual sample, which is used to reflect the memorization effect. Merits of the proposed historical statistic are two-fold: computationally efficient and robust to the randomness issue without the explicit label calibration in contrastive loss~\citep{chen2020simple}.  

\begin{figure*}[!t]
	\centering
	\includegraphics[width=1.00\textwidth]{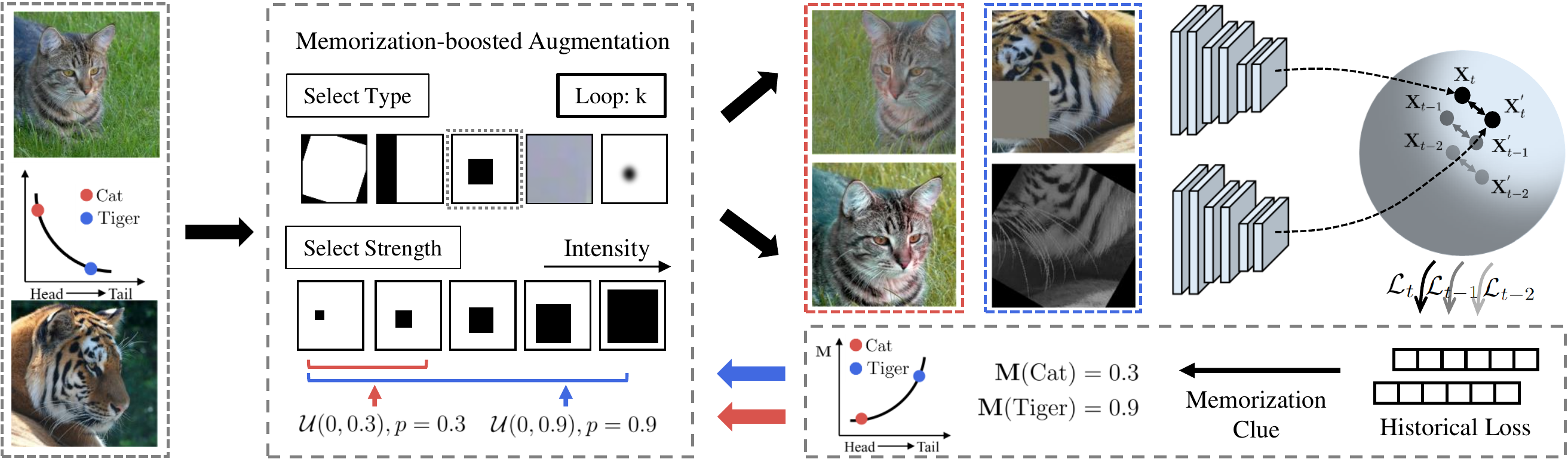}
	\hspace{-5mm}
    \caption{The illustration of Boosted Contrastive Learning. We trace the historical losses of each sample to find the clues about the memorization effect of DNNs, which then drives the augmentation strength to enhance the learning law on the tail samples. The ``head'' and ``tail'' indicators about the cat image and the tiger image are the exemplars and actually unknown during the training.}
    \label{fig:methods}
\end{figure*}
Besides, we boost the performance of contrastive learning on tail samples from the data perspective, \textit{i.e.,} construct the heavier information discrepancy between two views of the sample instead of the previous loss reweighting~\citep{lin2017focal,liu2021self} or the model pruning~\citep{jiang2021self}. According to the ``InfoMin Principle'' \citep{tian2020makes}, a good set of views are those that share the minimal information necessary to perform well at the downstream task. In this spirit, BCL dynamically constructs the information discrepancy between views to boost representation learning based on the memorization effect. Specifically, \method constructs the stronger information discrepancy between views to emphasize the importance of tail samples, while maintains the relative high correlation between views for head samples to avoid fitting to task-irrelevant noise. This allows our model to capture more task-relevant information from samples in long-tailed distribution. 

\subsection{Boosted Contrastive Learning}

In this section, we will present the formulation of the proposed Boosted Constrastive Learning, which leverages a momentum loss proxy to control the augmentation to affect the memorization effect of DNNs. Specifically, as tail samples tend to be learned slowly, they will be assigned with higher intensities of augmentation. Then, the model is driven to extract more information from the augmented views of tail samples for the better generalization.

Concretely, given a training sample $x_i$ on the long-tailed dataset, we denote its contrastive loss as $\mathcal{L}_i$ and  $\{\mathcal{L}_{i,0},\ldots,\mathcal{L}_{i,t},\ldots,\mathcal{L}_{i,T}\}$ traces a sequence of the loss values $\mathcal{L}_i$ in $T$ epochs. We then define the following moving-average momentum loss,
\begin{equation}\label{eq:momentum_loss}
    \mathcal{L}_{i,0}^m = \mathcal{L}_{i,0}, \ \ \mathcal{L}_{i,t}^m = \beta\mathcal{L}_{i,t-1}^m +(1-\beta)\mathcal{L}_{i,t}  \nonumber
\end{equation}
where $\beta$ is a hyper-parameter to control the degree smoothed by the historical losses. After the training in the $t$-th epoch through the above moving-average, we could acquire a set of the momentum losses for each sample as $\{\mathcal{L}_{0,t}^m,\ldots,\mathcal{L}_{i,t}^m,\ldots,\mathcal{L}_{N,t}^m\}$, where $N$ is the number of training samples in the dataset. Finally, we define the following normalization on the momentum losses,
\begin{equation}\label{eq:normalization}
    \mathbf{M}_{i,t} = \frac{1}{2}	\left(\frac{\mathcal{L}_{i,t}^m-\bar{\mathcal{L}}_{t}^m}{\ \ \ \ \ \ \max \left\{\left| \mathcal{L}_{i,t}^m-\bar{\mathcal{L}}_{t}^m\right|\right\}_{i=0,\dots,N}}+1\right)
\end{equation}
where $\bar{\mathcal{L}}_{t}^m$ is the average momentum loss at the $t$-th training epoch. By Eq.~\eqref{eq:normalization}, $\mathbf{M}_{i}$ is normalized to $[0,1]$ with the average value of $0.5$, which reflects the intensity of the memorization effect. To boost the contrastive learning, we use $\mathbf{M}_{i}$ as an indicator controlling the occurrence and strength of the augmentation. Specifically, we randomly selects $k$ types of augmentations from RandAugment~\citep{cubuk2020randaugment} and apply each augmentation with probability $\mathbf{M}_{i}$ and strength $[0,\mathbf{M}_{i}]$, respectively. 
For clarity, we assume augmentations defined in RandAugment as $\mathcal{A} = (A_{1},\ldots,A_{j},\ldots,A_{K})$, where $K$ denotes the amount of augmentations. In each step, only $k$ augmentations are applied~\citep{cubuk2020randaugment}. We formulate the memorization-boosted augmentation $\Psi (x_{i})$:
\begin{equation} \label{eq:aug}
\begin{aligned}
   &\Psi (x_{i}; \mathcal{A}, \mathbf{M}_{i}) = a_{1}(x_{i})\circ\ldots\circ a_{k}(x_{i}),  \\
   &a_j(x_i)=\begin{cases}
   A_{j}(x_i;\mathbf{M}_{i} \zeta) & u\sim \mathcal U(0,1)~\&~u < \mathbf{M}_{i} \\
    x_i &  \text{otherwise}\\
    \end{cases}
\end{aligned}    
\end{equation}
where $\zeta$ is sampled from the uniform distribution $\mathcal U(0,1)$ and $a_j(x_i)$ means we decide to keep $x_i$ unchanged or augment $x_i$ by $A_{j}(x_i;\mathbf{M}_{i} \zeta)$ based on whether $u$ is greater than $\mathbf{M}_{i}$. $A_{j}(x_i;\mathbf{M}_{i} \zeta)$ represents applying the $j$-th augmentation to $x_i$ with the strength $\mathbf{M}_{i} \zeta$, and $\circ$ is the function composition\footnote{\url{https://en.wikipedia.org/wiki/Function_composition}} operator, namely, sequentially applying the selected $k$ augmentations in $\mathcal{A}$. For simiplicity, we use $\Psi (x_{i})$ to represent $\Psi (x_{i}; \mathcal{A}, \mathbf{M}_{i})$ in this paper. Our boosted contrastive learning loss are formulated as follows.
\begin{equation}\label{eq:bcl}
\mathcal{L}_{\mathrm{BCL}}=\frac{1}{N} \sum_{i=1}^{N}-\log \frac{\exp \left(\frac{f(\Psi(x_{i}))^\top f(\Psi(x_{i}^{+}))}{\tau}\right)}{\sum_{x_{i}' \in X^{'}} \exp \left(\frac{f(\Psi(x_{i}))^\top f(\Psi(x_{i}'))}{\tau}\right)}
\end{equation}
where $X^{'}$ represents $X^{-}\cup \{x_i^+\}$ as Eq.~\eqref{eq:cl}. Intuitively, at a high level, \method can be understood as a curriculum learning method that adaptively assigns the appropriate augmentation strength for the individual sample according to the feedback from the memorization clues. Let $\theta$ denotes the model parameters and we have the following procedure
\begin{equation}
\begin{aligned}
   &\theta = \mathop{\arg\min}_{\theta} \mathcal{L}_{\mathrm{BCL}} \left( X, \Psi, \theta \right), \\
   & \Psi = \Psi (x; \mathcal{A}, \mathbf{M}), 
   \ \ \mathbf{M} = \mathrm{Normalize}\left(\mathcal{L}_{\mathrm{BCL}}^m\right) . 
   \nonumber
\end{aligned} 
\end{equation}
In this way,  \method continually depends on $\Psi$ to highlight the training samples to which DNNs show the poor memorization effect until its momentum loss $\mathcal{L}_{\mathrm{BCL}}^m$ degrades. By iteratively optimizing the model and building the memorization-boosted information discrepancy, we adaptively motivate model to learn “residual” information contained in tail samples. Note that, the form of $\mathcal{L}_{\mathrm{BCL}}$ can be flexibly replaced by the extensions from more self-supervised methods. In this paper, we mainly investigate two \method types, \emph{i.e.}, \methodnospace-I (Identity) and \methodnospace-D (Damaging). Specifically, \methodnospace-I means the plain \method in Eq.~\eqref{eq:bcl}, while \methodnospace-D is built on SDCLR and is formulated by the following equation,
\begin{align}\label{eq:bdcl}
\begin{split}
\mathcal{L}_{\mathrm{BCL-D}}=\frac{1}{N} \sum_{i=1}^{N}-\log \frac{\exp \left(\frac{f(\Psi(x_{i}))^\top g(\Psi(x_{i}^{+}))}{\tau}\right)}{\sum_{x_{i}' \in X^{'}} \exp \left(\frac{f(\Psi(x_{i}))^\top g(\Psi(x_{i}'))}{\tau}\right)}
\end{split}
\end{align}
where $g$ is the pruned version of $f$ as detailed in SDCLR~\citep{jiang2021self}. We illustrate \method in Figure~\ref{fig:methods} and summarize the complete procedure in Algorithm~\ref{alg:bcl}.

\begin{algorithm}[t!]
   \caption{Boosted Contrastive Learning (BCL)}
   \label{alg:bcl}
    {\bf Input:} dataset $\mathcal{X}$, the epoch number $T$, the weighting factor $\beta$, the number $k$ used in RandAugment, the whole augmentation set $\mathcal{A}$ ($K$ augmentation types)\\
    {\bf Output:} pretrained model parameter $\theta_{T}$\\
    {\bf Initialize:} model parameter $\theta_{0}$
\begin{algorithmic}[1]
    \IF{t $= 0$}
    \STATE Train model $\theta_{0}$ with Eq.~\eqref{eq:cl} and initialize $\mathcal{L}_{0}^m$, $\mathbf{M}_0$.
    \ENDIF
    \FOR{t $= 1$, $\dots$, $T-1$}
    \FOR{$x$ in $\mathcal{X}$}
    \STATE Select $k$ augmentations from the augmentation set $\mathcal{A}$ and construct augmented views $\Psi_{t} (x)$ according to $\mathbf{M}_{t-1}$ with Eq.~\eqref{eq:aug}. 
    \ENDFOR
    \STATE Train model $\theta_{t}$ with Eq.~\eqref{eq:bcl} or Eq.~\eqref{eq:bdcl} and obtain $\mathcal{L}_{t}$;
    \STATE Obtain $\mathcal{L}_{t}^m = \beta\mathcal{L}_{t-1}^m +(1-\beta)\mathcal{L}_{t}$ with stored $\mathcal{L}_{t-1}^m$;
    \STATE Update $\mathbf{M}_t \leftarrow \frac{1}{2}	\left(\frac{\mathcal{L}_{i,t}^m-\bar{\mathcal{L}}_{t}^m}{\ \ \ \   \max \left\{\left| \mathcal{L}_{i,t}^m-\bar{\mathcal{L}}_{t}^m\right|\right\}_{i=0,\dots,N}}+1\right)$;
 
    \ENDFOR
\end{algorithmic}
\end{algorithm}
\subsection{More Discussions on \methodnospace}

\textbf{Complexity.} The additional storage in BCL compared with that in the standard contrastive learning methods is the momentum loss. In Eq.~\eqref{eq:momentum_loss}, we only need to save a scalar $\mathcal{L}_{i,t-1}^m$ of the previous epoch for each sample. Therefore, its storage cost is as cheap as that of one label in the float type. 

\textbf{Compatibility.} \method does not require the specific model structures and thus it is compatible with many self-supervised learning methods in the recent years~\citep{chen2020simple,he2020momentum,byol,ermolov2021whitening,chen2021exploring}. Besides, it can be potentially adapted to enhance the representation learning under the supervised long-tailed learning setting in the form of pre-training or regularization for the representation learning of head and tail samples.

\textbf{Relation to loss re-weighting.} Loss re-weighting is an explicit way to enhance the learning of the specific samples by enlarging the importance of their losses. Previous attempts like Focal loss~\citep{lin2017focal} and SAM~\citep{liu2021self} belong to this case. In comparison, \method does not directly modify the loss, but captures the memorization clues to drive the construction of information discrepancy for the implicit re-weighting. In the following section, we will show that this actually is a more efficient way to bootstrap the long-tailed representation learning without label annotations.

\section{Experiments}

\subsection{Datasets and Baselines}

We conduct extensive experiments on three benchmark long-tailed datasets: CIFAR-100-LT~\cite{cao2019learning}, ImageNet-LT~\cite{liu2019large} and Places-LT~\cite{liu2019large} . 

\textbf{CIFAR-100-LT:} The original CIFAR-100 is a small-scale dataset composed of $32 \times 32$ images from 100 classes. For the long-tailed version, we use the same sampled subsets of CIFAR-100 as in \cite{jiang2021self}. The imbalace factor is defined by the number of the most frequent classes divided by the least frequent classes. Following~\cite{jiang2021self}, we set the imbalance factor as 100 and conduct experiments on five long-tailed splits to avoid randomness.

\textbf{ImageNet-LT:} ImageNet-LT~\cite{liu2019large} is a long-tailed version of ImageNet, which is down-sampled according to the Pareto distribution with the power value $\alpha=6$. It contains 115.8K images of 1000 categories, ranging from 1,280 to 5 in terms of the class cardinality.

\textbf{Places-LT:} Places~\cite{zhou2017places} is a large-scale scene-centric dataset and Places-LT is a long-tailed subset of Places following the Pareto distribution~\cite{liu2019large}. It contains 62,500 images in total from 365 categories, ranging from 4,980 to 5 under the class cardinality. 
\begin{table*} [!t]
\centering
\caption{Fine-grained analysis for various methods pre-trained on CIFAR-100-LT, ImageNet-LT and Places-LT. Many/Medium/Few corresponds to three partitions on the long-tailed data. Std is the standard deviation of the accuracies among Many/Medium/Few groups.} 

\resizebox{\linewidth}{!}{
\begin{small}
\begin{tabular}{  c c c c c c c c c c c c c c c}
\toprule
\multicolumn{1}{c}{} & \multicolumn{4}{c}{CIFAR-100-LT} & & \multicolumn{4}{c}{ImageNet-LT} & & \multicolumn{4}{c}{Places-LT} \\ 
 Methods  & Many & Medium & Few & Std & & Many & Medium & Few & Std & & Many & Medium & Few & Std   \\ \midrule
 SimCLR  & 48.70 & 46.81 & 44.02 & 2.36         & & 41.16  & 32.91  & 31.76 &5.13 & & 31.12 & 33.85 & 35.62 & 2.27 \\
 Focal  & 48.46 & 46.73 & 44.12 &\textbf{2.18}  & & 40.55 & 32.91 & 31.29 & 4.95 & & 30.18 & 31.56 & 33.32 & \underline{1.57} \\
 DnC  & \underline{54.00} & 46.68 & 45.65 &4.55             & & 29.54 & 19.62 & 18.38 & 6.12 & & 28.20 & 28.07 & 28.46 & \textbf{0.20}  \\
 SDCLR  & \underline{51.22}  & \underline{49.22} & \underline{45.85} & 2.71         & & \underline{41.24}  & \underline{33.62} & \underline{32.15} & \underline{4.88} & & \underline{32.08} & \underline{35.08} & \underline{35.94} & \underline{2.03} \\
 \midrule
 BCL-I  & 50.45 & \underline{48.23} & \underline{45.97} & \underline{2.24} & & \textbf{42.53}  & \textbf{35.66}  & \underline{33.93} & \underline{4.54} & & \underline{32.27} & \underline{34.96} & \textbf{38.03} & 2.88  \\
 BCL-D  & \textbf{53.98}  & \textbf{51.97}  & \textbf{49.52} & \underline{2.23}  
 & & \underline{41.92}  & \underline{35.29} & \textbf{34.07} & \textbf{4.22} & & \textbf{32.34} & \textbf{35.44} & \underline{37.75} & 2.71  \\

\bottomrule
 
\end{tabular}
\end{small}}
\label{tab:disjoint}
\end{table*}

\textbf{Baselines: }
To demonstrate the effectiveness of our method on benchmark datasets, we compare to many self-supervised methods related under long-tailed representation learning, including: (1) \emph{contrastive learning baseline}: SimCLR~\cite{chen2020simple}, (2) \emph{hard example mining}: Focal loss~\cite{lin2017focal}, (3) \emph{model ensemble}: DnC~\cite{tian2021divide}, (4) \emph{model damaging}: SDCLR~\cite{jiang2021self}. As mentioned before, BCL can be combined with any self-supervised learning architectures. Here, we term its combination with SimCLR as BCL-I and its combination with SDCLR as BCL-D, respectively.

\subsection{Implementation Details}
For all experiments, we use the SGD optimizer and the cosine annealing schedule. Similar to the backbone architecture and projection head proposed in~\cite{chen2020simple}, we use ResNet-18~\cite{he2016deep} as the backbone for experiments on CIFAR-100-LT and ResNet-50 on ImageNet-LT and Places-LT. The smoothing factor $\beta$ in the momentum loss Eq.~\eqref{eq:momentum_loss} is set as 0.97. Besides, we set $k=1$ for BCL-I and $k=2$ for BCL-D in the RandAugment. The whole augmentation set $\mathcal{A}$ is aligned with RandAugment where $K=16$. For the other pre-training settings, we follow~\cite{jiang2021self} and during evaluation, we leverage the way in~\cite{ermolov2021whitening}. Specifically, we train the classifier for 500 epochs and employ the learning rate decaying from $10^{-2}$ to $10^{-6}$. We use the Adam optimizer with the weight decay $5 \times 10^{-6}$.

We follow \cite{ermolov2021whitening} to conduct \emph{linear probing evaluation}, where a linear classifier is trained on top of the frozen pretrained backbone and the test accuracy is calculated to measure the representation quality. To eliminate the effect of long-tailed distribution in the fine-tuning stage, the classifier is trained on a balanced dataset. 
Specifically, we report the \emph{few-shot} performance of the classifier on basis of the pretrained representation. In the default case, we conduct 100-shot evaluation on CIFAR-100-LT, ImageNet-LT and Places-LT for performance evaluation. Meanwhile, we also implement the full-shot, 100-shot and 50-shot evaluation for abalation study on CIFAR-100-LT.

To visualize the fine-grained performance under the long-tailed setting, we divide each dataset to three partitions (\emph{Many-Medium-Few}).  Following~\cite{jiang2021self} on CIFAR-100-LT, the resulted partitions are Many (34 classes, 500 to 106 samples in the cardinal classes), Medium (33 classes, 105 to 20 samples in the cardinal classes) and Few (33 classes, 19 to 5 samples in the cardinal classes), respectively. As for the large-scale datasets ImageNet-LT and Places-LT, we follow~\cite{liu2019large} to divide each dataset into Many (over 100 samples), Medium (100 to 20 samples) and Few (under 20 samples). The average accuracy and the standard deviation are computed among three groups.

\begin{table}[t]
\caption{The overall performance of various methods pre-trained on CIFAR-100-LT, ImageNet-LT and Places-LT with 100-shot.}
\label{tab:imagenet-places}

\begin{center}
\begin{small}
\begin{tabular}{lcccc}
\toprule
Methods & CIFAR-100-LT & ImageNet-LT & Places-LT  \\
\midrule
SimCLR & 46.53 & 35.93 & 33.22  \\
Focal & 46.46 & 35.63  & 31.41  \\
DnC   & \underline{48.53}  & 23.27  & 28.19  \\
SDCLR & \underline{48.79} & \underline{36.35} & \underline{34.17} \\ 
\midrule
BCL-I & 48.24 & \textbf{38.07} & \underline{34.59} \\
BCL-D & \textbf{51.84} & \underline{37.68} & \textbf{34.78}  \\

\bottomrule
\end{tabular}
\end{small}
\end{center}
\end{table}
\subsection{Performance Evaluation}

\textbf{Overall performance.} In Table~\ref{tab:imagenet-places}, we summarize the performance of different methods on three long-tailed datasets. According to the results, BCL-I and BCL-D significantly improve the few-shot
performance by 1.71\% and 3.05\% on SimCLR and SDCLR on CIFAR-100-LT. On large-scale datasets ImageNet-LT and Places-LT, compared with SimCLR, SDCLR only improves the few-shot accuracy by 0.42\% and 0.95\%. However, our methods maintain a consistent gain over other self-supervised methods and specifically, BCL-I achieves comparable performance with BCL-D and outperforms SDCLR by 1.72\% on ImageNet-LT.

\textbf{Fine-grained analysis.} In Table~\ref{tab:disjoint}, we visualize the merit of BCL in the fine-grained perspective. According to the results on CIFAR-100-LT, ImageNet-LT and Places-LT, we can see that \method achieves the new state-of-the-art performance on each partition across different benchmark datasets. For example, compared with SDCLR, BCL-D improves Many, Medium and Few accuracy by 2.77$\%$, 2.75$\%$ and 3.87$\%$ on CIFAR-100-LT, respectively. We also apply standard deviation(Std) of average accuracy on each partition to measure the representation balancedness. As shown in Table~\ref{tab:disjoint}, we see that our methods reduce Std by a considerable margin of 0.4$-$0.7 on CIFAR-100-LT and ImageNet-LT. 
Note that, the results on Places-LT differ from the former datasets as the performance of three groups shows a reverse trend on the long-tailed distribution. Nevertheless, an interesting observation is that BCL-I still significantly improves Few accuracy by 2.09\% while maintain at Many(0.19\%) compared with SDCLR. The results confirm that \method can boost the performance on tail classes and potentially handle the more complicated real-world data distribution.

\begin{figure}[!t]
    \centering
    \includegraphics[width=1\linewidth]{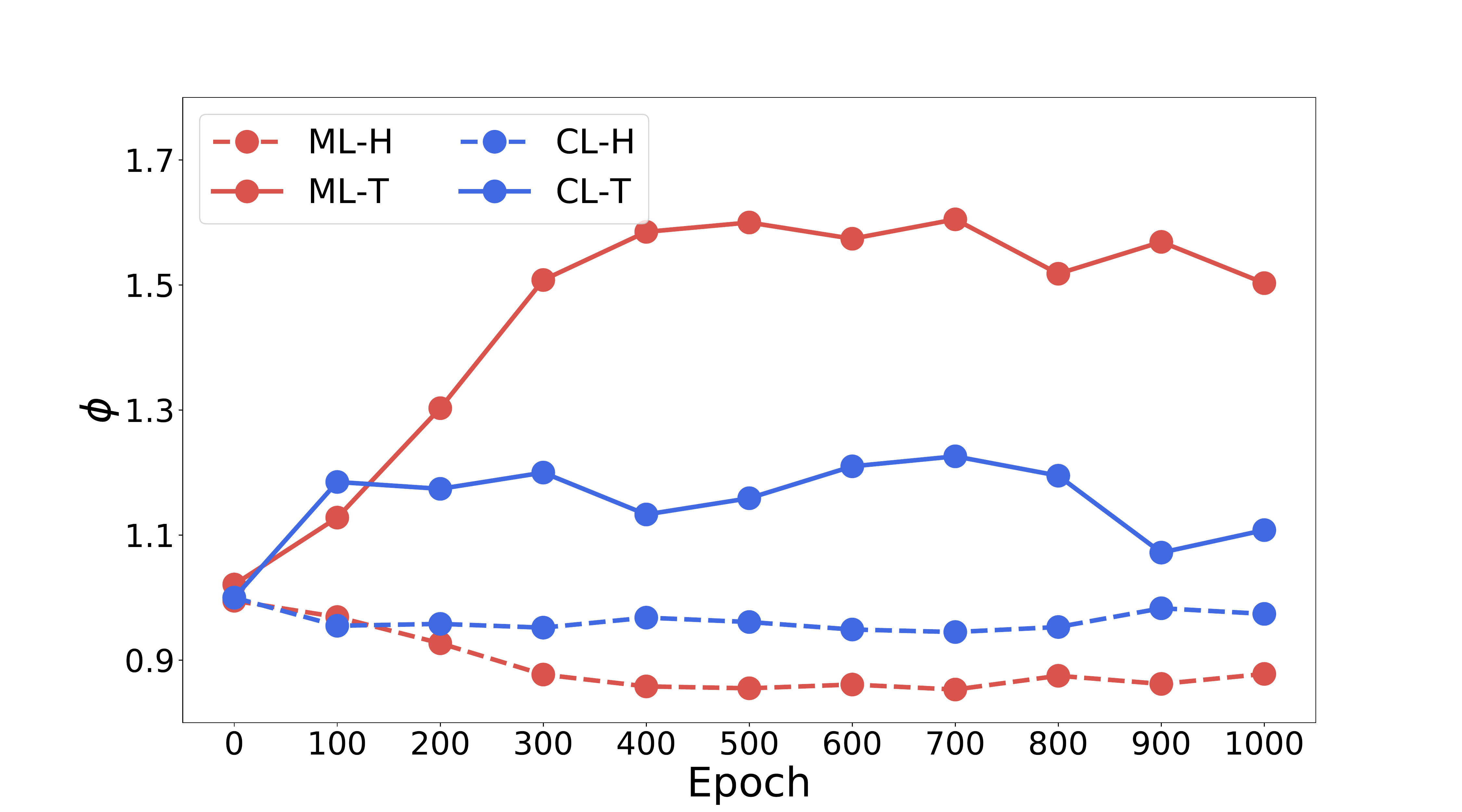}
    \caption{Long-tailed sample discovery with our momentum loss (ML) and the conventional contrastive loss (CL) under different training epochs on CIFAR-100-LT. $\phi$ means the proportion of head or tail classes in the top 10\% large-loss samples of the dataset. }
    \label{fig:tail_detector}
\end{figure}

\textbf{Long-tailed sample discovery.} We use ground-truth labels to validate the tail detection of the momentum loss mechanism in Eq.~\eqref{eq:momentum_loss}. First, we pre-train SimCLR and store the loss value of the training sample in each epoch. We then calculate the momentum loss and choose the training samples that have top-10$\%$ highest loss. To mitigate the effect of the group size, we apply the correlation metric in~\cite{jiang2021improving} and divide the train dataset into head (Major) and tail (Medium, Few). Specifically, the metric is defined as: 
\begin{equation}
\phi=\frac{\mathcal{G} \cap \mathcal{X}_{l}}{\mathcal{G} \cap \mathcal{X}}, \ \ \ \ \mathcal{X}_{l} = \mathop{\arg\max}_{\mathcal{X}^{\prime}:\left|\mathcal{X}^{\prime}\right| \geq r|\mathcal{X}| }\mathcal{L}\left( \mathcal{X}^{\prime} \right) \nonumber
\end{equation}
where $\mathcal{G}$ denotes the target group, $\mathcal{X}_{l}$ represents the subset of large-loss samples and $r$ represents the threshold ratio. We set $r = 0.1$ and compare the proposed $\mathcal{L}^m_{\mathrm{CL}}$ with $\mathcal{L}_{\mathrm{CL}}$.

As shown in Figure~\ref{fig:tail_detector}, we can find that more tail samples are extracted by the proposed momentum loss compared those by the standard contrastive loss. Meanwhile, we find that momentum loss serves as a reliable tail detector only except the early stage of training process in Figure~\ref{fig:tail_detector}. As the momentum loss is built on the historical information, a long-term observation will yield a more stable estimation.

\begin{table}[t!]
\caption{The classification accuracy of supervised learning with self-supervised pre-training on CIFAR-100-LT and ImageNet-LT. }
\label{tab:finetune}
\centering
\resizebox{\linewidth}{!}{
\begin{tabular}{  c c c c c c  c c  }
\toprule
\multirow{2}{*}{Dataset}   & \multirow{2}{*}{CE} & \multicolumn{6}{c}{CE with the following model initialization}   \\ 
  &   & CL & Focal & DnC & SDCLR & BCL-I & BCL-D   \\ \midrule
CIFAR-100-LT  & \colorbox{gray!30}{41.7} & 44.4 & 44.4 & 44.4   & \underline{44.6}  &  \textbf{45.1} & \textbf{45.4} \\
ImageNet-LT  & \colorbox{gray!30}{41.6}  & 45.5  & 45.4 & 42.2  &  \underline{45.9}  & \textbf{46.9} & \textbf{46.4} \\ \midrule
\multirow{2}{*}{Dataset}   & \multirow{2}{*}{cRT} & \multicolumn{6}{c}{cRT with the following model initialization}   \\ 
  &   & CL & Focal & DnC & SDCLR & BCL-I & BCL-D   \\ \midrule
CIFAR-100-LT  & \colorbox{gray!30}{44.1} & 48.9 & 48.7 &  48.6  & \underline{49.8}  & \textbf{49.9} & \textbf{50.0} \\
ImageNet-LT  & \colorbox{gray!30}{46.7}  & \underline{47.5}  & 47.3  & 43.5  &  47.3  & \textbf{48.4}  & \textbf{48.1}  \\ \midrule
\multirow{2}{*}{Dataset}   & \multirow{2}{*}{LA} & \multicolumn{6}{c}{LA with the following model initialization}   \\ 
  &   & CL & Focal & DnC & SDCLR & BCL-I & BCL-D   \\ \midrule
CIFAR-100-LT  & \colorbox{gray!30}{45.7} & 50.1 & 49.5  & 49.7   & \underline{50.4}  & \textbf{50.8}  & \textbf{50.5} \\
ImageNet-LT  & \colorbox{gray!30}{47.4}  & \underline{48.6}  & 48.4  & 45.6  &  48.2  & \textbf{49.7} & \textbf{49.1} \\
\bottomrule
\end{tabular}}
\end{table}

\begin{table*} [!t]
\centering
\caption{The linear probing performance of all methods on CUB, Cars, Aircrafts, Dogs and NABirds. We pretrain the backbone ResNet-50 on ImageNet-LT under different methods, and then transfer to these datasets for the linear probing evaluation. The top-1 and top-5 accuracies are reported by computing the highest and top-5 highest predictions to match the ground-truth labels.} 

\resizebox{\linewidth}{!}{
\begin{small}
{ 
\begin{tabular}{ c c c c c c c c c c c c c }
\toprule
\multicolumn{1}{c}{} & \multicolumn{2}{c}{CUB} & \multicolumn{2}{c}{Cars} & \multicolumn{2}{c}{Aircrafts} & \multicolumn{2}{c}{Dogs} & \multicolumn{2}{c}{NABirds} & \multicolumn{2}{c}{All} \\

Methods & Top-1 & Top-5 & Top-1 & Top-5& Top-1 & Top-5& Top-1 & Top-5 & Top-1 & Top-5 & Top-1 & Top-5 \\
\midrule
SimCLR  & \underline{29.62} &\underline{57.35} &21.45 &44.93 &30.48 &57.01 & 46.67 &79.22  &\underline{16.52}  &\underline{37.61}  &  28.95 & 55.22  \\
Focal   & \underline{29.08} & 56.89 & 21.40 & 44.35 & 30.99 & 57.64 & 46.59 & 78.14 & 16.31 & 36.97 & 28.87 & 54.80 \\
DnC   & 16.97 & 40.90 & 8.15 & 23.79 & 13.71 & 33.18 & 29.83 & 61.92 & 8.44 & 22.75 & 15.42  & 36.51 \\
SDCLR   & 28.98 & 57.27 & \underline{22.10} & \underline{46.13} & \underline{31.05} & \underline{58.18} & \underline{46.69} & \underline{78.82} & 16.17 & 37.10 & \underline{29.00}  & \underline{55.50} \\
\midrule
BCL-I & \textbf{30.00} & \textbf{58.08} & \underline{23.67} & \underline{49.16} & \underline{32.37} &\underline{60.31} & \textbf{48.61} &\textbf{79.99}  & \textbf{17.42} &\textbf{38.96}  & \underline{30.41} & \underline{57.30} \\
BCL-D & 28.79 &\underline{57.37} & \textbf{25.90} & \textbf{51.34} & \textbf{34.95} &\textbf{62.77} & \underline{47.49} &\underline{78.86}  & \underline{16.41} &\underline{37.24}  & \textbf{30.71} & \textbf{57.51}  \\

\bottomrule
\end{tabular}
}
\end{small}}

\label{tab:downstream}
\end{table*}
\subsection{On Transferability for Downstream Tasks}

\textbf{Downstream supervised long-tailed classification.} Self-supervised pre-training is proved to be useful for learning more generalizable representations by label-agnostic model initialization~\citep{yang2020rethinking}. All the self-supervised long-tailed baselines can be regarded as the pre-training methods that are compatible with supervised algorithms. In order to validate the effectiveness of \method on downstream supervised long-tailed tasks, we use the pre-trained self-supervised models to initialize the supervised model backbone and then finetune all parameters. Specifically, we evaluate and compare 3 representative long-tailed methods: Cross Entropy, cRT~\citep{kang2019decoupling} and Logit Adjustment~\citep{menonlong} with 6 self-supervised initialization methods on CIFAR-100-LT and ImageNet-LT. The results of the finetuning experiment are summarized in Table~\ref{tab:finetune}, showing that initialization with self-supervised models always helps improve over the standard baseline, and BCL outperforms all other self-supervised pre-training methods. This indicates the potential merits of BCL to further boost the supervised long-tailed representation learning.  

\textbf{Downstream fine-grained classification.}  In order to validate the representation transferability of our memorization-boosted augmentation, we conduct experiments on various downstream fine-grained datasets: Caltech-UCSD Birds (CUB200)~\citep{wah2011caltech}, Stanford Cars~\citep{krause20133d}, Aircrafts~\citep{maji2013fine}, Stanford Dogs~\citep{khosla2011novel}, NABirds~\citep{van2015building}. The training and testing images of these datasets roughly range from 10k to 50k. Meanwhile, these datasets include five distinct categorizes, from birds to cars, where the intrinsic property of the data distribution varies. We first pre-train the model on ImageNet-LT and then conduct the linear probing evaluation on these target datasets individually. 

In Table~\ref{tab:downstream}, we present the transfer results on various downstream tasks. According to the table, we can see that our methods consistently surpass other methods with a considerable margin in all cases. Specifically, our methods significantly improve the best Top-1 accuracy by 3.80\%, 3.90\% and 1.92\% on Stanford Cars, Aircrafts and Dogs, and by 0.38\% and 0.90\% on the other two bird datasets, CUB and NABirds. Overall, BCL-D on average improves Top-1 and Top-5 accuracy by 1.71$\%$ and 2.01$\%$ on five target datasets.  This confirms our intuition that there is discarded transferable information for tail samples, which is effectively extracted by \methodnospace. Tracing out distinct mutual information for head or tail samples, \method encourages to learn more generalizable and robust representation on the long-tailed dataset compared to the baselines from the loss and model perspectives.

\subsection{Ablation Study}
\textbf{On augmentation components.} In Table~\ref{tab:component}, we conduct various experiments to investigate the effect of individual augmentation components in $\mathcal{A}$ of BCL. Specifically, we set the augmentation number $k=1$ and additionally add each component to the sampled subset of augmentations. In this way, the monitored component dominants to construct the information discrepancy in the training stage. We then evaluate the effect of each component by computing the difference on the linear probing accuracy compared with \emph{Identity} augmentation (i.e. $k=1$) on CIFAR-100-LT. As shown in Table~\ref{tab:component}, we can see that the geometric-related augmentations are more helpful for representation learning. In particular, \emph{ShearX}, \emph{ShearY} and \emph{Cutout} significantly improve the linear probing accuracy by $0.69\%$, $0.90\%$ and $0.82\%$, respectively. However, some color-related augmentations lead to the degeneration of the linear probing accuracy except \emph{Posterize}, \emph{Sharpness} and \emph{Brightness}. Intuitively, the color distortion augmentations in standard setting might be enough for contrastive learning methods, while some geometric-related semantics can further be captured by \methodnospace.

\begin{figure*}[!t]
	\centering
	\includegraphics[width=0.35\textwidth]{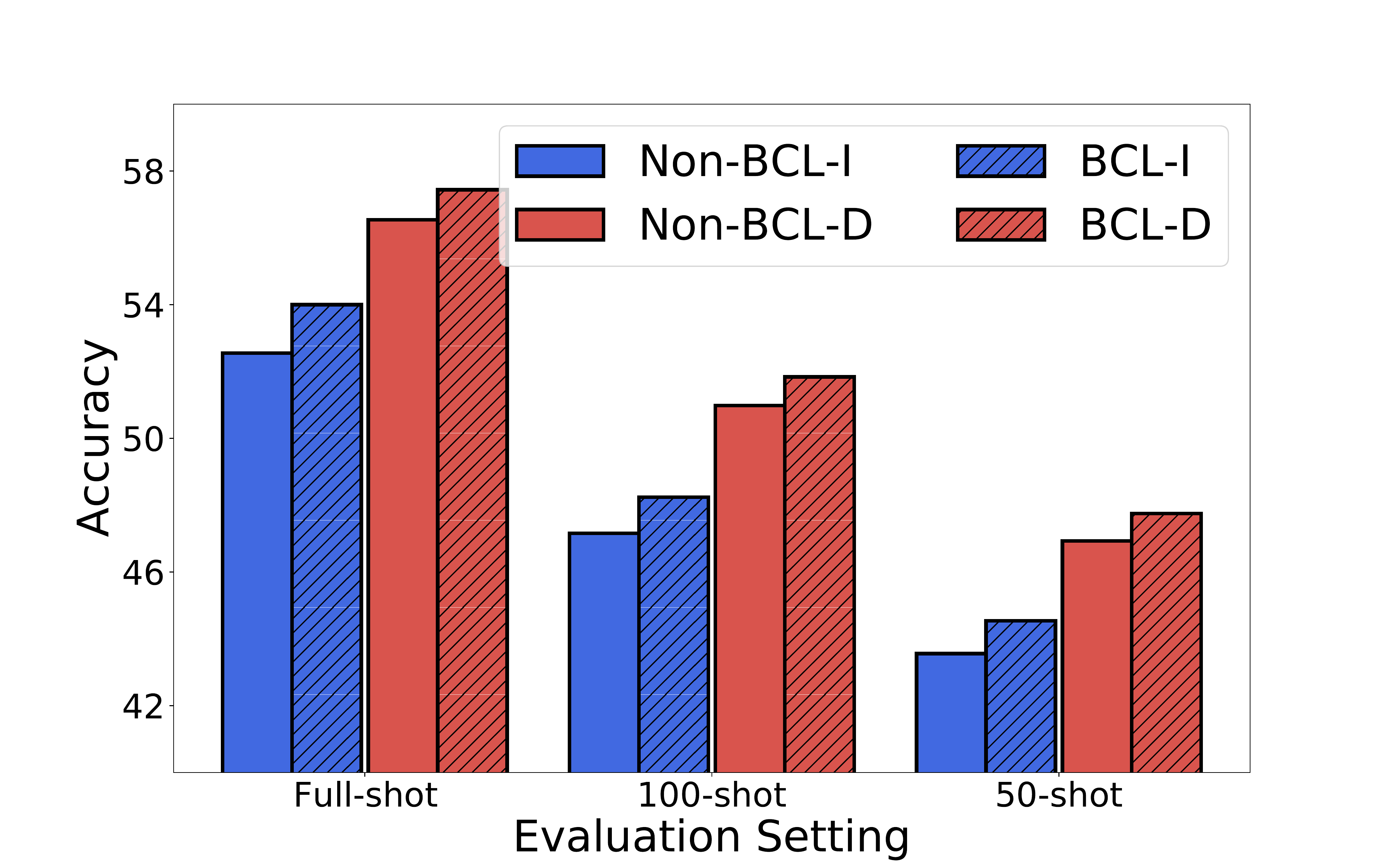}
	\hspace{-7mm}
	\includegraphics[width=0.35\textwidth]{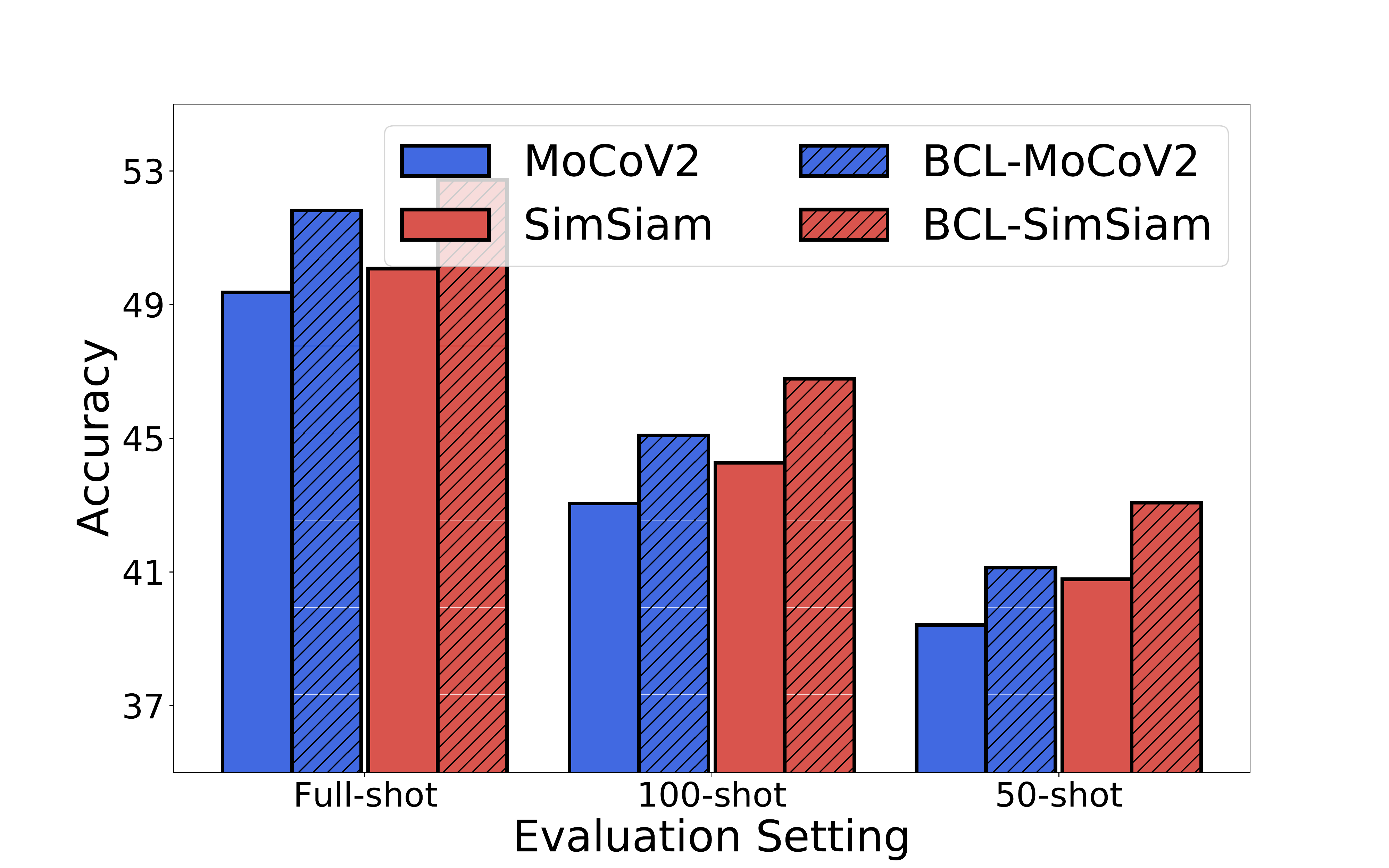}
	\hspace{-7mm}
	\includegraphics[width=0.35\textwidth]{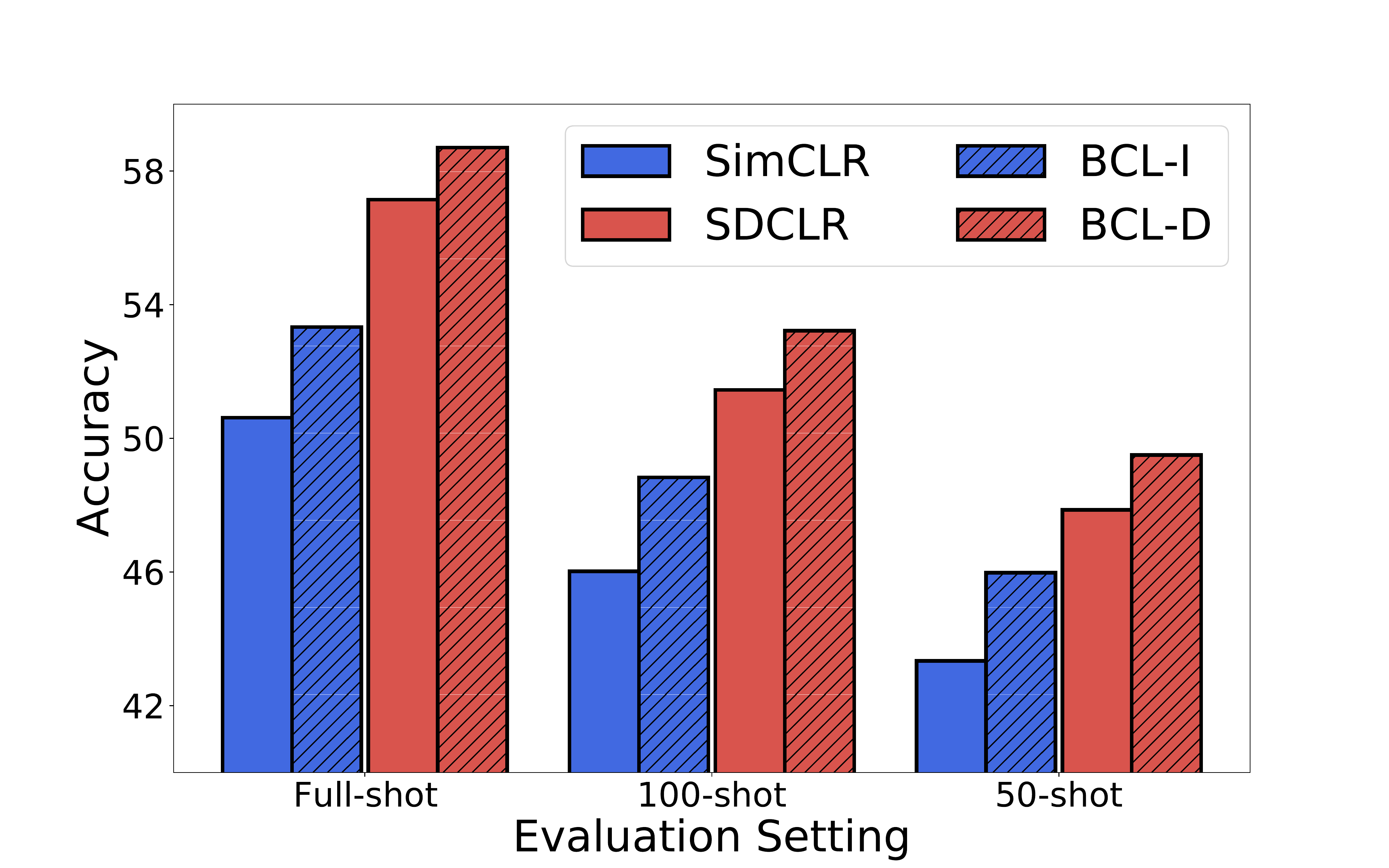} 
    \caption{(Left) Linear probing evaluation under different shots for BCL and Non-BCL (without the memorization guidance) pre-trained on CIFAR-100-LT. (Middle) Linear probing evaluation under different shots for MoCoV2 and SimSiam pre-trained on CIFAR-100-LT. (Right) Linear probing evaluation under different shots for BCL pre-trained on CIFAR-100, compared with SimCLR and SDCLR.}
    \label{fig:ablation}
\end{figure*}

\textbf{Augmentation \textit{w vs. w/o} the memorization guidance.} To study the importance of the memorization guidance for the augmentation, we compare to RandAugment combined with SimCLR and SDCLR. For fair comparison, we fix the strength of augmentation in RandAugment. Note that, non-BCL means adopting strong and uniform augmenation to all samples in the dataset. Therefore, the performance bias from the augmentation is decoupled in these experiments. As shown in the left panel of Figure~\ref{fig:ablation}, we can see that BCL consistently outperforms Non-BCL on linear probing evaluation under different shot on CIFAR-100-LT. Specifically, BCL-I and BCL-D improve full-shot performance by 1.45\% and 0.9\%, compared with Non-BCL-I and Non-BCL-D. The results confirm the effectiveness of the tailness detection mechanism and memorization-boosted design in \methodnospace.

\textbf{Other contrastive learning backbones.} We extend our BCL to another two representative contrastive learning methods: MoCoV2~\cite{he2020momentum,chen2020improved}, SimSiam~\cite{chen2021exploring}, as shown in the middle panel of Figure~\ref{fig:ablation}. From the results, we can see that BCL maintains a consistent gain over MoCoV2 and SimSiam. The improvements show that BCL is orthogonal to the current self-supervised learning methods in long-tailed scenarios.

\begin{table}[!t]
\caption{Improvement of linear probing performance on additionally adopting each component relative to that with \emph{Identity} augmentation for BCL on CIFAR-100-LT. $\Delta(\%)$ means the relative gain.}
\label{tab:component}
\begin{center}
\begin{small}
\begin{tabular}{lccc}
\toprule
Component & $\Delta(\%)$ & Component & $\Delta(\%)$ \\
\midrule

Identity & 0.00  &  Equalize & -1.28 \\
ShearX   & 0.69  & Solarize  & -2.69  \\
ShearY   & \textbf{0.90}      & Posterize  & 0.59 \\
TranslateX  & 0.44  & Contrast  & -0.2 \\
TranslateY  & 0.37  & Color  & -0.53 \\
Rotate      & 0.13  & Brightness  & -0.08 \\
Cutout  & 0.82  & Sharpness   & -0.06  \\
Invert      & \textbf{-4.38}  & AutoContrast  & -0.96 \\

\bottomrule
\end{tabular}
\end{small}
\end{center}
\end{table}

\textbf{Performance on balanced datasets.} Following \cite{jiang2021self}, we also validate our BCL on balanced subsets of CIFAR-100 to explore whether BCL can benefit “implicit imbalancedness” \emph{w.r.t.}, atypical samples or sampling bias on balanced data. Results are shown in the right panel of Figure~\ref{fig:ablation}.  Similarly, BCL boosts the linear probing performance by $1.56\%$, $1.77\%$ and $1.64\%$ under different evaluations.

\begin{figure}[!t]
	\centering
	\includegraphics[width=0.25\textwidth]{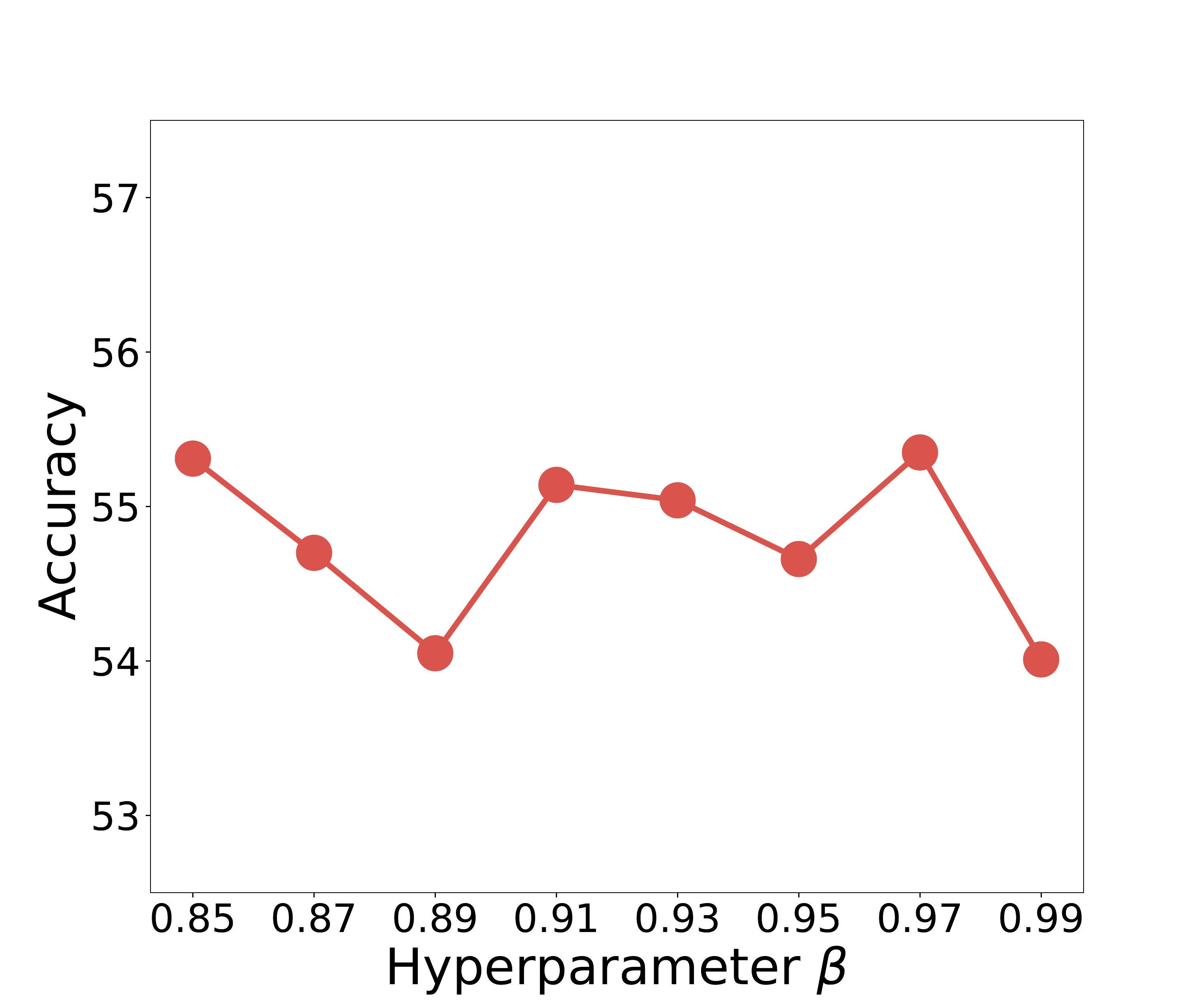}
	\hspace{-5mm}
	\includegraphics[width=0.25\textwidth]{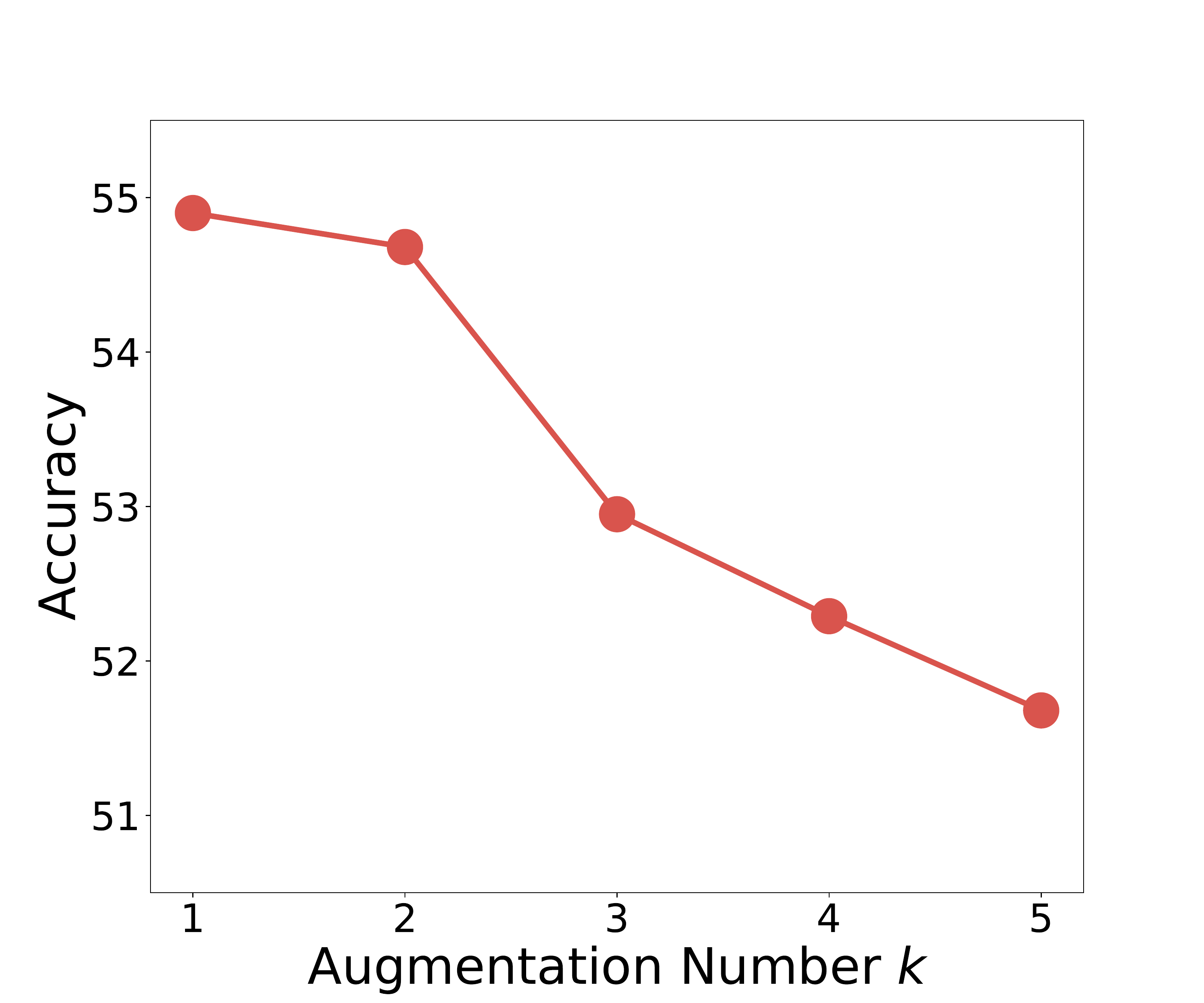} 
    \caption{(Left) Linear probing performance under different $\beta$ for BCL-I on CIFAR-100-LT. (Right) Linear probing performance with different $k$ for BCL-I on CIFAR-100-LT.}
    \label{fig:beta_aug_k}
\end{figure}

\textbf{Impact of $\beta$ in the momentum loss Eq.~\eqref{eq:momentum_loss}} In the left panel of Figure~\ref{fig:beta_aug_k}, we conduct several experiments with different $\beta$ value to validate the stability of BCL.  We compare different $\beta$ in a high range (0.85-0.99) as the longer observations of the memorization effect are preferred to construct a reliable tail discovery. From the curve, we can see that BCL is mostly promising as the performance fluctuates a little.

\textbf{Different augmentation number $k$.} In the right panel of Figure~\ref{fig:beta_aug_k}, we validate BCL by training with different numbers of augmentations sampled from RandAugment. We can see that BCL achieves the appealing results with $k=1,2$ but degenerates at settings with the higher augmentation number $k$. Specifically, our method achieves 54.90\% and 54.68\% when adopting $k=1, 2$ for the RandAugment, and 52.95\%, 52.29\%, 51.68\% for $k=3, 4, 5$, respectively. The performance difference reaches 3.22\% between $k=1$ and $k=5$. We trace several augmented views and find that they are extremely distorted with limited information available when adopting $k=5$ for RandAugment.  We conjecture that too strong augmentation may lead to too much information loss and it becomes hard for \method to encode the important details to the representation. On the other hand, a smaller $k$ is also preferred due to the small computational cost.

\section{Conclusion}

In this paper, we propose a novel Boosted Contrastive Learning~(BCL) method for the representation learning under the long-tailed data distribution. It leverages the clues of memorization effect in the historical training losses to automatically construct the information discrepancy for head and tail samples, which then drives contrastive learning to pay more attention to the tail samples. Different from previous methods that builds in the perspective of the loss or the model, \method is essentially from the data perspective and orthogonal to the early explorations. Through extensive experiments, we demonstrate the effectiveness of \method under different settings. In the future, we will extend \method to more challenging long-tailed data like iNaturalist and explore the properties of the tail samples in more practical scenarios.

\section*{Acknowledgement}
This work is partially supported by the National Key R\&D Program of China (No. 2019YFB1804304), 111 plan (No. BP0719010), STCSM (No. 18DZ2270700, No. 21DZ1100100), and State Key Laboratory of UHD Video and Audio Production and Presentation. BH was supported by the RGC Early Career Scheme No.~22200720, NSFC Young Scientists Fund No.~62006202, and Guangdong Basic and Applied Basic Research Foundation No.~2022A1515011652.

\bibliography{main}

\begin{thebibliography}{49}
\providecommand{\natexlab}[1]{#1}
\providecommand{\url}[1]{\texttt{#1}}
\expandafter\ifx\csname urlstyle\endcsname\relax
  \providecommand{\doi}[1]{doi: #1}\else
  \providecommand{\doi}{doi: \begingroup \urlstyle{rm}\Url}\fi

\bibitem[Arpit et~al.(2017)Arpit, Jastrz{\k{e}}bski, Ballas, Krueger, Bengio,
  Kanwal, Maharaj, Fischer, Courville, Bengio, et~al.]{arpit2017closer}
Arpit, D., Jastrz{\k{e}}bski, S., Ballas, N., Krueger, D., Bengio, E., Kanwal,
  M.~S., Maharaj, T., Fischer, A., Courville, A., Bengio, Y., et~al.
\newblock A closer look at memorization in deep networks.
\newblock In \emph{International Conference on Machine Learning}, pp.\
  233--242. PMLR, 2017.

\bibitem[Brown et~al.(2020)Brown, Mann, Ryder, Subbiah, Kaplan, Dhariwal,
  Neelakantan, Shyam, Sastry, Askell, et~al.]{brown2020language}
Brown, T., Mann, B., Ryder, N., Subbiah, M., Kaplan, J.~D., Dhariwal, P.,
  Neelakantan, A., Shyam, P., Sastry, G., Askell, A., et~al.
\newblock Language models are few-shot learners.
\newblock \emph{Advances in Neural Information Processing Systems},
  33:\penalty0 1877--1901, 2020.

\bibitem[Cao et~al.(2019)Cao, Wei, Gaidon, Arechiga, and Ma]{cao2019learning}
Cao, K., Wei, C., Gaidon, A., Arechiga, N., and Ma, T.
\newblock Learning imbalanced datasets with label-distribution-aware margin
  loss.
\newblock \emph{Advances in Neural Information Processing Systems},
  32:\penalty0 1567--1578, 2019.

\bibitem[Caron et~al.(2020)Caron, Misra, Mairal, Goyal, Bojanowski, and
  Joulin]{caron2020unsupervised}
Caron, M., Misra, I., Mairal, J., Goyal, P., Bojanowski, P., and Joulin, A.
\newblock Unsupervised learning of visual features by contrasting cluster
  assignments.
\newblock \emph{Advances in Neural Information Processing Systems},
  33:\penalty0 9912--9924, 2020.

\bibitem[Chen et~al.(2020{\natexlab{a}})Chen, Kornblith, Norouzi, and
  Hinton]{chen2020simple}
Chen, T., Kornblith, S., Norouzi, M., and Hinton, G.
\newblock A simple framework for contrastive learning of visual
  representations.
\newblock In \emph{International Conference on Machine Learning}, pp.\
  1597--1607. PMLR, 2020{\natexlab{a}}.

\bibitem[Chen \& He(2021)Chen and He]{chen2021exploring}
Chen, X. and He, K.
\newblock Exploring simple siamese representation learning.
\newblock In \emph{Proceedings of the IEEE/CVF Conference on Computer Vision
  and Pattern Recognition}, pp.\  15750--15758, 2021.

\bibitem[Chen et~al.(2020{\natexlab{b}})Chen, Fan, Girshick, and
  He]{chen2020improved}
Chen, X., Fan, H., Girshick, R., and He, K.
\newblock Improved baselines with momentum contrastive learning.
\newblock \emph{arXiv preprint arXiv:2003.04297}, 2020{\natexlab{b}}.

\bibitem[Cubuk et~al.(2020)Cubuk, Zoph, Shlens, and Le]{cubuk2020randaugment}
Cubuk, E.~D., Zoph, B., Shlens, J., and Le, Q.~V.
\newblock Randaugment: Practical automated data augmentation with a reduced
  search space.
\newblock In \emph{Proceedings of the IEEE/CVF Conference on Computer Vision
  and Pattern Recognition Workshops}, pp.\  702--703, 2020.

\bibitem[Cui et~al.(2021)Cui, Zhong, Liu, Yu, and Jia]{cui2021parametric}
Cui, J., Zhong, Z., Liu, S., Yu, B., and Jia, J.
\newblock Parametric contrastive learning.
\newblock In \emph{Proceedings of the IEEE/CVF International Conference on
  Computer Vision}, pp.\  715--724, 2021.

\bibitem[Cuturi(2013)]{cuturi2013sinkhorn}
Cuturi, M.
\newblock Sinkhorn distances: Lightspeed computation of optimal transport.
\newblock \emph{Advances in Neural Information Processing Systems},
  26:\penalty0 2292--2300, 2013.

\bibitem[Deng et~al.(2009)Deng, Dong, Socher, Li, Li, and
  Fei-Fei]{deng2009imagenet}
Deng, J., Dong, W., Socher, R., Li, L.-J., Li, K., and Fei-Fei, L.
\newblock Imagenet: A large-scale hierarchical image database.
\newblock In \emph{2009 IEEE Conference on Computer Vision and Pattern
  Recognition}, pp.\  248--255. Ieee, 2009.

\bibitem[Doersch et~al.(2015)Doersch, Gupta, and
  Efros]{doersch2015unsupervised}
Doersch, C., Gupta, A., and Efros, A.~A.
\newblock Unsupervised visual representation learning by context prediction.
\newblock In \emph{Proceedings of the IEEE International Conference on Computer
  Vision}, pp.\  1422--1430, 2015.

\bibitem[Ermolov et~al.(2021)Ermolov, Siarohin, Sangineto, and
  Sebe]{ermolov2021whitening}
Ermolov, A., Siarohin, A., Sangineto, E., and Sebe, N.
\newblock Whitening for self-supervised representation learning.
\newblock In \emph{International Conference on Machine Learning}, pp.\
  3015--3024. PMLR, 2021.

\bibitem[Feldman(2020)]{feldman2020does}
Feldman, V.
\newblock Does learning require memorization? a short tale about a long tail.
\newblock In \emph{Proceedings of the 52nd Annual ACM SIGACT Symposium on
  Theory of Computing}, pp.\  954--959, 2020.

\bibitem[Grill et~al.(2020)Grill, Strub, Altch\'{e}, Tallec, Richemond,
  Buchatskaya, Doersch, Avila~Pires, Guo, Gheshlaghi~Azar, Piot, kavukcuoglu,
  Munos, and Valko]{byol}
Grill, J.-B., Strub, F., Altch\'{e}, F., Tallec, C., Richemond, P.,
  Buchatskaya, E., Doersch, C., Avila~Pires, B., Guo, Z., Gheshlaghi~Azar, M.,
  Piot, B., kavukcuoglu, k., Munos, R., and Valko, M.
\newblock Bootstrap your own latent - a new approach to self-supervised
  learning.
\newblock In \emph{Advances in Neural Information Processing Systems},
  volume~33, pp.\  21271--21284, 2020.

\bibitem[Han et~al.(2018)Han, Yao, Yu, Niu, Xu, Hu, Tsang, and
  Sugiyama]{han2018co}
Han, B., Yao, Q., Yu, X., Niu, G., Xu, M., Hu, W., Tsang, I.~W., and Sugiyama,
  M.
\newblock Co-teaching: Robust training of deep neural networks with extremely
  noisy labels.
\newblock In \emph{Advances in Neural Information Processing Systems}, 2018.

\bibitem[He et~al.(2016)He, Zhang, Ren, and Sun]{he2016deep}
He, K., Zhang, X., Ren, S., and Sun, J.
\newblock Deep residual learning for image recognition.
\newblock In \emph{Proceedings of the IEEE Conference on Computer Vision and
  Pattern Recognition}, pp.\  770--778, 2016.

\bibitem[He et~al.(2020)He, Fan, Wu, Xie, and Girshick]{he2020momentum}
He, K., Fan, H., Wu, Y., Xie, S., and Girshick, R.
\newblock Momentum contrast for unsupervised visual representation learning.
\newblock In \emph{Proceedings of the IEEE/CVF Conference on Computer Vision
  and Pattern Recognition}, pp.\  9729--9738, 2020.

\bibitem[Jiang et~al.(2018)Jiang, Zhou, Leung, Li, and
  Fei-Fei]{jiang2018mentornet}
Jiang, L., Zhou, Z., Leung, T., Li, L.-J., and Fei-Fei, L.
\newblock Mentornet: Learning data-driven curriculum for very deep neural
  networks on corrupted labels.
\newblock In \emph{International Conference on Machine Learning}, pp.\
  2304--2313. PMLR, 2018.

\bibitem[Jiang et~al.(2021{\natexlab{a}})Jiang, Chen, Chen, and
  Wang]{jiang2021improving}
Jiang, Z., Chen, T., Chen, T., and Wang, Z.
\newblock Improving contrastive learning on imbalanced data via open-world
  sampling.
\newblock \emph{Advances in Neural Information Processing Systems}, 34,
  2021{\natexlab{a}}.

\bibitem[Jiang et~al.(2021{\natexlab{b}})Jiang, Chen, Mortazavi, and
  Wang]{jiang2021self}
Jiang, Z., Chen, T., Mortazavi, B.~J., and Wang, Z.
\newblock Self-damaging contrastive learning.
\newblock In \emph{International Conference on Machine Learning}, pp.\
  4927--4939. PMLR, 2021{\natexlab{b}}.

\bibitem[Jiang et~al.(2021{\natexlab{c}})Jiang, Zhang, Talwar, and
  Mozer]{jiang2020characterizing}
Jiang, Z., Zhang, C., Talwar, K., and Mozer, M.~C.
\newblock Characterizing structural regularities of labeled data in
  overparameterized models.
\newblock In \emph{International Conference on Machine Learning}, volume 139,
  pp.\  5034--5044. PMLR, 2021{\natexlab{c}}.

\bibitem[Kang et~al.(2019)Kang, Xie, Rohrbach, Yan, Gordo, Feng, and
  Kalantidis]{kang2019decoupling}
Kang, B., Xie, S., Rohrbach, M., Yan, Z., Gordo, A., Feng, J., and Kalantidis,
  Y.
\newblock Decoupling representation and classifier for long-tailed recognition.
\newblock In \emph{International Conference on Learning Representations}, 2019.

\bibitem[Kang et~al.(2020)Kang, Li, Xie, Yuan, and Feng]{kang2020exploring}
Kang, B., Li, Y., Xie, S., Yuan, Z., and Feng, J.
\newblock Exploring balanced feature spaces for representation learning.
\newblock In \emph{International Conference on Learning Representations}, 2020.

\bibitem[Khosla et~al.(2011)Khosla, Jayadevaprakash, Yao, and
  Li]{khosla2011novel}
Khosla, A., Jayadevaprakash, N., Yao, B., and Li, F.-F.
\newblock Novel dataset for fine-grained image categorization: Stanford dogs.
\newblock In \emph{Proc. CVPR Workshop on Fine-Grained Visual Categorization
  (FGVC)}, volume~2. Citeseer, 2011.

\bibitem[Krause et~al.(2013)Krause, Stark, Deng, and Fei-Fei]{krause20133d}
Krause, J., Stark, M., Deng, J., and Fei-Fei, L.
\newblock 3d object representations for fine-grained categorization.
\newblock In \emph{Proceedings of the IEEE International Conference on Computer
  Vision workshops}, pp.\  554--561, 2013.

\bibitem[Lan et~al.(2019)Lan, Chen, Goodman, Gimpel, Sharma, and
  Soricut]{lan2019albert}
Lan, Z., Chen, M., Goodman, S., Gimpel, K., Sharma, P., and Soricut, R.
\newblock Albert: A lite bert for self-supervised learning of language
  representations.
\newblock \emph{International Conference on Learning Representation}, 2019.

\bibitem[Li et~al.(2021)Li, Cao, Yuan, Fan, Yang, Feris, Indyk, and
  Katabi]{li2021targeted}
Li, T., Cao, P., Yuan, Y., Fan, L., Yang, Y., Feris, R., Indyk, P., and Katabi,
  D.
\newblock Targeted supervised contrastive learning for long-tailed recognition.
\newblock \emph{arXiv preprint arXiv:2111.13998}, 2021.

\bibitem[Lin et~al.(2017)Lin, Goyal, Girshick, He, and
  Doll{\'a}r]{lin2017focal}
Lin, T.-Y., Goyal, P., Girshick, R., He, K., and Doll{\'a}r, P.
\newblock Focal loss for dense object detection.
\newblock In \emph{Proceedings of the IEEE International Conference on Computer
  Vision}, pp.\  2980--2988, 2017.

\bibitem[Liu et~al.(2021)Liu, HaoChen, Gaidon, and Ma]{liu2021self}
Liu, H., HaoChen, J.~Z., Gaidon, A., and Ma, T.
\newblock Self-supervised learning is more robust to dataset imbalance.
\newblock \emph{arXiv preprint arXiv:2110.05025}, 2021.

\bibitem[Liu et~al.(2019)Liu, Miao, Zhan, Wang, Gong, and Yu]{liu2019large}
Liu, Z., Miao, Z., Zhan, X., Wang, J., Gong, B., and Yu, S.~X.
\newblock Large-scale long-tailed recognition in an open world.
\newblock In \emph{Proceedings of the IEEE/CVF Conference on Computer Vision
  and Pattern Recognition}, pp.\  2537--2546, 2019.

\bibitem[Maji et~al.(2013)Maji, Rahtu, Kannala, Blaschko, and
  Vedaldi]{maji2013fine}
Maji, S., Rahtu, E., Kannala, J., Blaschko, M., and Vedaldi, A.
\newblock Fine-grained visual classification of aircraft.
\newblock \emph{arXiv preprint arXiv:1306.5151}, 2013.

\bibitem[Menon et~al.(2021)Menon, Jayasumana, Jain, Veit, Kumar, and
  Rawat]{menonlong}
Menon, A.~K., Jayasumana, S., Jain, H., Veit, A., Kumar, S., and Rawat, A.~S.
\newblock Long-tail learning via logit adjustment.
\newblock In \emph{International Conference on Learning Representations}, 2021.

\bibitem[Reed(2001)]{reed2001pareto}
Reed, W.~J.
\newblock The pareto, zipf and other power laws.
\newblock \emph{Economics letters}, 74\penalty0 (1):\penalty0 15--19, 2001.

\bibitem[Ren et~al.(2018)Ren, Zeng, Yang, and Urtasun]{ren2018learning}
Ren, M., Zeng, W., Yang, B., and Urtasun, R.
\newblock Learning to reweight examples for robust deep learning.
\newblock In \emph{International Conference on Machine Learning}, pp.\
  4334--4343. PMLR, 2018.

\bibitem[Tian et~al.(2020)Tian, Sun, Poole, Krishnan, Schmid, and
  Isola]{tian2020makes}
Tian, Y., Sun, C., Poole, B., Krishnan, D., Schmid, C., and Isola, P.
\newblock What makes for good views for contrastive learning?
\newblock In \emph{Advances in Neural Information Processing Systems}, pp.\
  6827--6839, 2020.

\bibitem[Tian et~al.(2021)Tian, Henaff, and van~den Oord]{tian2021divide}
Tian, Y., Henaff, O.~J., and van~den Oord, A.
\newblock Divide and contrast: Self-supervised learning from uncurated data.
\newblock In \emph{Proceedings of the IEEE/CVF International Conference on
  Computer Vision}, pp.\  10063--10074, 2021.

\bibitem[Van~Horn et~al.(2015)Van~Horn, Branson, Farrell, Haber, Barry,
  Ipeirotis, Perona, and Belongie]{van2015building}
Van~Horn, G., Branson, S., Farrell, R., Haber, S., Barry, J., Ipeirotis, P.,
  Perona, P., and Belongie, S.
\newblock Building a bird recognition app and large scale dataset with citizen
  scientists: The fine print in fine-grained dataset collection.
\newblock In \emph{Proceedings of the IEEE Conference on Computer Vision and
  Pattern Recognition}, pp.\  595--604, 2015.

\bibitem[Van~Horn et~al.(2018)Van~Horn, Mac~Aodha, Song, Cui, Sun, Shepard,
  Adam, Perona, and Belongie]{van2018inaturalist}
Van~Horn, G., Mac~Aodha, O., Song, Y., Cui, Y., Sun, C., Shepard, A., Adam, H.,
  Perona, P., and Belongie, S.
\newblock The inaturalist species classification and detection dataset.
\newblock In \emph{Proceedings of the IEEE Conference on Computer Vision and
  Pattern Recognition}, pp.\  8769--8778, 2018.

\bibitem[Wah et~al.(2011)Wah, Branson, Welinder, Perona, and
  Belongie]{wah2011caltech}
Wah, C., Branson, S., Welinder, P., Perona, P., and Belongie, S.
\newblock The caltech-ucsd birds-200-2011 dataset.
\newblock 2011.

\bibitem[Wang \& Isola(2020)Wang and Isola]{wang2020understanding}
Wang, T. and Isola, P.
\newblock Understanding contrastive representation learning through alignment
  and uniformity on the hypersphere.
\newblock In \emph{International Conference on Machine Learning}, pp.\
  9929--9939. PMLR, 2020.

\bibitem[Wang \& Gupta(2015)Wang and Gupta]{wang2015unsupervised}
Wang, X. and Gupta, A.
\newblock Unsupervised learning of visual representations using videos.
\newblock In \emph{Proceedings of the IEEE International Conference on Computer
  Vision}, pp.\  2794--2802, 2015.

\bibitem[Yang \& Xu(2020)Yang and Xu]{yang2020rethinking}
Yang, Y. and Xu, Z.
\newblock Rethinking the value of labels for improving class-imbalanced
  learning.
\newblock In \emph{NeurIPS}, 2020.

\bibitem[Yu et~al.(2019)Yu, Han, Yao, Niu, Tsang, and Sugiyama]{yu2019does}
Yu, X., Han, B., Yao, J., Niu, G., Tsang, I., and Sugiyama, M.
\newblock How does disagreement help generalization against label corruption?
\newblock In \emph{International Conference on Machine Learning}, pp.\
  7164--7173. PMLR, 2019.

\bibitem[Zhang et~al.(2017)Zhang, Bengio, Hardt, Recht, and
  Vinyals]{zhang2017understanding}
Zhang, C., Bengio, S., Hardt, M., Recht, B., and Vinyals, O.
\newblock Understanding deep learning requires rethinking generalization.
\newblock In \emph{International Conference on Learning Representations}, 2017.

\bibitem[Zhang et~al.(2021)Zhang, Kang, Hooi, Yan, and Feng]{zhang2021deep}
Zhang, Y., Kang, B., Hooi, B., Yan, S., and Feng, J.
\newblock Deep long-tailed learning: A survey.
\newblock \emph{arXiv preprint arXiv:2110.04596}, 2021.

\bibitem[Zheng et~al.(2019)Zheng, Yao, Zhang, Tsang, and
  Wang]{zheng2019understanding}
Zheng, H., Yao, J., Zhang, Y., Tsang, I.~W., and Wang, J.
\newblock Understanding vaes in fisher-shannon plane.
\newblock In \emph{Proceedings of the AAAI Conference on Artificial
  Intelligence}, volume~33, pp.\  5917--5924, 2019.

\bibitem[Zheng et~al.(2021)Zheng, Chen, Yao, Yang, Li, Zhang, Zhang, Tsang,
  Zhou, and Zhou]{zheng2021contrastive}
Zheng, H., Chen, X., Yao, J., Yang, H., Li, C., Zhang, Y., Zhang, H., Tsang,
  I., Zhou, J., and Zhou, M.
\newblock Contrastive attraction and contrastive repulsion for representation
  learning.
\newblock \emph{arXiv preprint arXiv:2105.03746}, 2021.

\bibitem[Zhou et~al.(2017)Zhou, Lapedriza, Khosla, Oliva, and
  Torralba]{zhou2017places}
Zhou, B., Lapedriza, A., Khosla, A., Oliva, A., and Torralba, A.
\newblock Places: A 10 million image database for scene recognition.
\newblock \emph{IEEE Transactions on Pattern Analysis and Machine
  Intelligence}, 40\penalty0 (6):\penalty0 1452--1464, 2017.

\end{thebibliography}
\bibliographystyle{icml2022}

\end{document}